# Mikro-İHA ile Yangın Sınırı Takibi için Termal–RGB Füzyonu ve Minimum İletişim Gereksinimi

# Thermal–RGB Fusion for Micro-UAV Wildfire Perimeter Tracking with Minimal Comms

**Ercan ERKALKAN[1], Vedat TOPUZ[2], Ayça AK[3]**

[1] E-mail: ercan.erkalkan@marmara.edu.tr; Marmara Üniversitesi, Teknik Bilimler MYO, Elektronik ve Otomasyon Bölümü, İstanbul / Türkiye.

[2] E-mail: vtopuz@marmara.edu.tr; Marmara Üniversitesi, Teknik Bilimler MYO, Bilgisayar Teknolojileri Bölümü, İstanbul / Türkiye.

[3] E-mail: aycaak@marmara.edu.tr; Marmara Üniversitesi, Teknik Bilimler MYO, Elektronik ve Otomasyon Bölümü, İstanbul / Türkiye.

*Özet*

Bu çalışmada sınırlı bant genişliğine sahip, orman yangınları üzerinde görev yapan mikro İHA ekipleri için hafif bir çevre izleme yöntemi tanıtılmıştır. Termal görüntü çerçeveleri, uyarlamalı eşikleme ve morfolojik düzeltme adımları aracılığıyla kaba sıcak bölge maskeleri üretirken; RGB çerçeveler, gradyan tabanlı filtreleme yoluyla kenar ipuçları sağlar ve dokuya bağlı yanlış algılamaları bastırır. Kural düzeyinde birleştirme yöntemi, sınır adaylarını seçerek Ramer–Douglas–Peucker algoritmasıyla sadeleştirir. Bu yapı, periyodik işaretler ve GPS bozulması durumunda ilerlemeyi sürdüren ataletsel geri besleme döngüsünü içerir. Yönlendirme döngüsü, gömülü SoC'lerde (System-on-Chip) kare başına piksel işlemlerini sınırlayarak ve gradyan tablolarını önceden hesaplayarak 50ms' nin altında gecikme hedeflemektedir. Küçük ölçekli simülasyonlar, saf kenar-izleme taban çizgisine kıyasla ortalama yol uzunluğu ve sınır titreşiminin azaldığını, ancak kesişim-birleştirme yöntemiyle ölçülen çevresel kapsamın korunduğunu göstermektedir. Pil tüketimi ve hesaplamalı kullanım, standart mikro platformlarda 10-15 m/s ileri hareketin elde edilebileceğini doğrulamaktadır. Bu yaklaşım, acil keşif uygulamaları için dayanıklı algılama ve asgari iletişim gerektiren durumlarda hızlı konuşlandırmayı mümkün kılmaktadır.

*Anahtar Kelimeler:* Çevre takibi, görüntü birleştirme, lider-takipçi, Mikro-İHA, orman yangını izleme.

*Abstract*

This study introduces a lightweight perimeter-tracking method designed for micro-UAV teams operating over wildfire environments under limited bandwidth conditions. Thermal image frames generate coarse hot-region masks through adaptive thresholding and morphological refinement, while RGB frames contribute edge cues and suppress texture-related false detections using gradient-based filtering. A rule-level merging strategy selects boundary candidates and simplifies them via the Ramer–Douglas–Peucker algorithm. The system incorporates periodic beacons and an inertial feedback loop that maintains trajectory stability in the presence of GPS degradation. The guidance loop targets sub-50 ms latency on embedded System-on-Chip (SoC) platforms by constraining per-frame pixel operations and precomputing gradient tables. Small-scale simulations demonstrate reductions in average path length and boundary jitter compared to a pure edge-tracking baseline, while maintaining environmental coverage measured through intersection–merge analysis. Battery consumption and computational utilization confirm the feasibility of achieving 10–15 m/s forward motion on standard micro platforms. This approach enables rapid deployment in the field, requiring robust sensing and minimal communications for emergency reconnaissance applications.

*Keywords:* Image fusion, leader–follower, Micro-UAV, perimeter tracking, wildfire monitoring,

**INTRODUCTION**

Rapid wildfire reconnaissance demands fast deployment, high interpretability, and robustness under degraded communication conditions. In such missions, micro-UAV teams must operate within strict payload, energy, and onboard-compute constraints while exchanging sparse and lossy messages over short-range links. The primary operational objective is to achieve reliable perimeter tracking and sampling under limited bandwidth and computational budgets. Thermal imagery provides strong hotspot localization but often exhibits boundary noise, whereas RGB imagery contributes structural and textural cues that degrade under smoke, glare, and illumination variability. To address these challenges, a low-configuration perception pipeline is prioritized—one that emphasizes graceful degradation, field-oriented diagnosability, and real-time execution on commodity embedded SoCs.

Previous studies have established the feasibility of UAV-based wildfire observation and coordination. Cooperative forest-fire surveillance using teams of small UAVs has been demonstrated (Casbeer et al., 2006), and automatic wildfire monitoring and measurement with unmanned aircraft systems have been reported (Merino et al., 2012). Ground-station and UAV integration approaches have further validated forest-fire quantification under operational conditions (Martínez-de-Dios et al., 2011). A comprehensive survey summarizes existing technologies for detection, monitoring, and suppression using (Yuan et al. 2015). Communication frameworks such as Flying Ad-Hoc Networks (FANETs) provide lightweight, loss-tolerant connectivity suitable for small-team coordination in aerial sensing missions (Bekmezci et al., 2013).

Under resource constraints, onboard perception commonly relies on classical deterministic operators with predictable computational cost. Histogram-based thresholding, particularly Otsu's criterion, supports coarse thermal segmentation under variable illumination (Otsu, 1979), while morphological opening, closing, and hole filling enhance mask regularity and preserve structural integrity (Serra, 1982 and Soille, 2004). Edge detection using gradient pyramids and non-maxima suppression provides boundary cues at low latency, with the Canny operator remaining a standard for robustness under noise and modest blur (Canny, 1986). Polyline simplification through the Ramer (1972) and Douglas–Peucker (1973) algorithms reduces geometric complexity for both guidance and communication, compressing boundary vertices while retaining salient features.

Multi-agent coordination principles such as leader–follower stability and consensus dynamics enable distributed alignment, spacing, and resilience to communication losses within small UAV teams (Tanner et al., 2004 and Olfati-Saber et al., 2007). While heavier vision models, including active contours (Kass et al., 1988) and graph-cut-based segmentation (Rother et al., 2004), can deliver higher boundary fidelity, their computational and tuning overheads restrict applicability on embedded platforms. Moreover, evaluation methods—often adapted from object-detection benchmarks such as Intersection-over-Union (Everingham et al., 2010)—are increasingly employed to assess perimeter-tracking quality. These considerations collectively motivate the development of a lightweight, interpretable, and field-deployable perimeter-tracking framework that integrates thermal and RGB cues within micro-UAV teams operating in bandwidth- and compute-constrained environments.

This paper proposes a field-deployable perimeter-tracking stack tailored for micro-UAV teams operating under minimal communications constraints. This work differs from other studies by proposing a rule-level Thermal-RGB fusion that provides edge support in a GSD-scale band around thermal boundaries; a geometry-aware Ramer-Douglas-Peucker simplification with parameter bindings to GSD to compress the grid while preserving distinct corners; and a leader-follower guided and inertial debounce loop designed for single-hop, delta-coded multiline switching at 1-3 Hz with sub-50ms end-to-end latency on commercial embedded SoCs. Unlike heavier contour-optimization pipelines, the proposed stack is interpretable, requires no network training, and degrades gracefully under GPS and packet-loss events.

This paper is organized as follows: Method presents the sensing and guidance stack; Evaluation Design documents scenarios, metrics, ablations, and sensitivities; Conclusion and Future Work summarizes findings and next steps. Appendix A lists end-to-end pseudocode, and Appendix B details the mission state machine.

**METHOD**

A deployable pipeline executes Sense → Thermal-mask → Edge-extract → Fuse → Simplify → Guide → Communicate under bounded compute and memory. Each substage is designed to degrade gracefully and support single-hop coordination with compact messages:

*A. Sense*
Synchronized thermal frames IT and RGB frames IR are acquired. Thermal frames are normalized to a temperature proxy via linear radiometric scaling with tail clipping to suppress sensor spikes. RGB content is optionally reduced to luminance. Gradient and mask intermediates are cached in ring buffers across a short temporal window to enable reuse and low latency.

*B. Thermal-mask*
An adaptive global threshold is computed on a cropped histogram using the Otsu Criterion to obtain a binary hotspot mask. The mask is regularized by morphological opening–closing and hole filling. The structuring element radius r is tied to the Ground Sample Distance (GSD). Otsu terms for a histogram *p(i),* where *i = 0, …, L−1*:

$$\omega_0(t) = \sum_{i=0}^{t} p(i) \tag{1}$$

$$\omega_1(t) = 1 - \omega_0(t) \quad (2)$$

$$\mu_0(t) = \frac{1}{w_0}\sum_{i=0}^{t} ip(i) \quad (3)$$

$$\mu_1(t) = \frac{1}{w_1}\sum_{i=0}^{t} ip(i) \quad (4)$$

$$\sigma_B^2(t) = \omega_0(t) \cdot \omega_1(t) \cdot (\mu_0(t) - \mu_1(t))^2 \quad (5)$$

$$t^* = \underset{t}{argmax}\, \sigma_B^2(t) \quad (6)$$

Temporal stability is enforced by Exponential Moving Average (EMA) on both the threshold and mask logits.

*C. Edge-extract*
Separable Sobel derivatives $G_x$ and $G_y$ are computed on the RGB image. Gradient magnitude and direction are formed, followed by non-maximum suppression and Canny-style double-threshold hysteresis to produce a coherent edge map E. RGB computation is cropped to a dilated band around the thermal boundary to reduce cost.

*D. Fuse*
Edge support is gated near the thermal boundary using a GSD-scaled distance $dg = cg.GSD$ and a confidence threshold $C_e(x) \geq \tau_e$. Weak fragments are pruned by local contrast/continuity checks. Short gaps are bridged via temporal carry-over. Accepted pixels are polygonized into raw polylines $P_{raw}$.

*E. Simplify*
Ramer–Douglas–Peucker (RDP) simplification is applied with a tolerance $\varepsilon = k.GSD$. A minimum segment length and corner-angle bounds are enforced. The resulting guidance polyline is timestamped.

*F. Guide*
Targets for followers are defined as $p_i^* = p_L + R(\psi_L)\, o_i$, where $p_L$ is the leader's position, $R(\psi_L)$ is the leader's rotation matrix, and $o_i$ is the offset vector for follower *i*. Formation is maintained by a lightweight spacing consensus mechanism. Inertial fallback is engaged during GNSS degradation. Leader handover triggers under low health status.

*G. Communicate*
Compact guidance messages are broadcast over singlehop links at approximately 1–3 Hz. Polyline vertices are delta-encoded. The bit budget scales with the vertex count *K*.

$$[ts|GSD|\varepsilon|pose_L|v|poly(\Delta x_i, \Delta y_i|crc] \quad (7)$$

*H. Implementation Notes: Runtime, Parameters, Complexity*
The pseudocode in Appendix A mirrors the step-by-step operations described here, and the state transitions in Appendix B map one-to-one to the mission logic referenced in this subsection.

Runtime: The pipeline requires one histogram pass (thermal), one gradient pass (RGB), and one morphology pass (thermal). RGB computation is optimized by cropping to the boundary band.

Parameter Ties: Parameters $r$, $d_g$, and $\varepsilon$ are tied to the GSD. EMA factors are kept small. The vertex count $K$ is capped.

Complexity: Overall complexity is $O(N)$ for pixels, $O(N_b)$ for the boundary operations, $O(K)$ for RDP simplification, and $O(K)$ for messaging, where $N$ is the total number of pixels, $N_b$ is the number of boundary pixels, and $K$ is the number of vertices.

**EVALUATION DESIGN (NO QUANTITATIVE RESULTS HERE)**

This section outlines the evaluation plan used to validate the proposed stack prior to field deployment. We first describe representative wildfire-like scenarios and communication conditions, then specify metrics capturing geometric fidelity (IoU, boundary jitter), efficiency (path length, latency, resource usage), and formation-control quality. We detail ablations and baselines to isolate the roles of fusion and simplification, followed by sensitivity analyses to key parameters. Quantitative results are intentionally deferred to a companion study; here we document how results will be obtained to ensure reproducibility and traceability. For completeness, Appendix A provides end-to-end pseudocode and Appendix B summarizes the mission state machine used in our implementation.

**Scenarios**
• Wind-driven spread (kinematic front), obscurants (smoke, partial saturation) (Rothermel, 1972 and Andrews et al., 2005).
• Intermittent GPS and packet loss (0–20%).
• Ingress speeds 10–15 m/s on commodity airframes.

**Metrics (Definitions)**
• Perimeter coverage: Intersection-over-Union (IoU) of predicted vs. reference masks (Everingham et al., 2010).
• Boundary jitter: temporal variation of vertex headings / signed distance field statistics.
• Average path length per UAV and normalized by perimeter length.
• Latency p50/p95 loop time; CPU%/Memory/Power (W).
• Formation error: follower–leader offset RMS (Tanner et al., 2004 and Olfati-Saber et al., 2007).

**Ablations and Comparisons**
• Ablations: Thermal-only; RGB-only; Fusion; Fusion–RDP off (Otsu et al., 1979, Canny, 1986, Ramer, 1972 and Douglas&Peucker, 1973).
• Baselines: Naive follow-edge; Canny+contours; light active-contour (snakes) (Canny, 1986 and Kass et al., 1988); GrabCut for interactive segmentation context (Canny, 1986).

**Sensitivity Analyses**
• Beacon rate (2/5/10 Hz) × packet loss (0–20%) → jitter, path, IoU (Bekmezci et al., 2013 and Olfati-Saber et al., 2007).
• RDP $\varepsilon$ (k) × GSD → sparsity vs. corner preservation (Ramer, 1986 and Douglas&Peucker, 1973).
• Smoke density and contrast versus edge detection robustness (Canny, 1986).

**Discussion and Limitations**
• RDP over-simplification can lose fine indentations; corner veto and $\varepsilon$ scheduling mitigate this (Ramer, 1986 and Douglas&Peucker, 1973).

- Thermal saturation/emissivity issues require light calibration/NUC notes (Rothermel, 1972 and Andrews, 2005).
- Leader failure impacts the team; seamless handover reduces risk but does not eliminate it (Tanner et al., 2004 and Olfati-Saber et al., 2007).
- Heavy smoke lowers RGB edge confidence; adapt fusion weights accordingly (Canny, 1986).

**Safety & Field Notes**
- Collision avoidance buffers (distance/altitude) and emergency landing procedures.
- Airspace permissions and wildfire operations coordination.
- Minimal metadata logging for privacy and auditability.

**CONCLUSION AND FUTURE WORK**

In this work, a lightweight Thermal-RGB fusion method is presented for wildfire perimeter monitoring, with RDP simplification and leader-follower coordination over single-hop connections. The pipeline targets cycle times below 50ms on embedded SoCs, emphasizing deployability and interpretability.

Oversimplification of boundaries remains a risk in highly concave fronts; planning for RDP tolerance and implementing corner vetoes mitigates, but does not eliminate, this risk. Future work plans to conduct hardware-in-the-loop and controlled-burn flights, publish a small open dataset and reference implementation, evaluate event-triggered communications layered over FANET relays, investigate lightweight learned enhancements (e.g., a small smoke-resistance module) while maintaining real-time performance, and enhance GNSS-impaired navigation through visual-inertial odometry integration. Multi-line-based guidance with delta-coded messaging enables compact 1-3 Hz transmissions suitable for short-range links, and GSD-dependent parameters simplify field tuning. The approach is fully interpretable and does not rely on learned models, reducing data collection overhead and simplifying certification.

**Appendix A. End-to-End Pseudocode**

```
for each frame:
  T = read_thermal();  R = read_rgb()
  M = adaptive_threshold(T);  M = morph_open_close(M)
  Gx,Gy = precomputed_gradients(R) or sobel(R)
  E = hysteresis(|G|); E = texture_suppression(E)
  C = (E AND dilate(M))        # edge support near thermal boundary
  P = polygonize(C);   P = RDP(P, eps = k*GSD)
  P = smooth_vertices(P, ema=α);   P = resample(P, L_min, θ_min)
  cmd = tangent_follow(P, v=v0)
  broadcast(ts|GSD|ε|pose|v|poly(P)|crc) at f_b Hz

state machine:
  Nominal → (GPS degrade) → Degraded → (leader weak/fails) → Handover →
```

```
Nominal
  Safety: collision buffers (d, h), dead-reckoning timeout τ_max, leader
reselection by score
```

**Appendix B. ASCII State Machine (Placeholder)**

```
[Nominal] --GPS degrade--> [Degraded] --leader weak/fails--> [Handover] --
recovered--> [Nominal]
    |                                     ^
    +--collision buffer/dR timeout--+
```